%% file: main.tex
\documentclass[10pt,twocolumn,letterpaper]{article}

\usepackage[pagenumbers]{cvpr} 

\usepackage{graphicx}
\usepackage{amsmath}
\usepackage{amssymb}
\usepackage{booktabs}

\usepackage{comment}
\usepackage{color}

\usepackage{booktabs}
\usepackage{algorithm}
\usepackage{algorithmic}
\usepackage{multirow}
\usepackage{tabularx}
\usepackage{verbatim}
\usepackage{subcaption}
\usepackage{pifont}

\usepackage[pagebackref=true,breaklinks=true,colorlinks,bookmarks=false]{hyperref}

\usepackage[capitalize]{cleveref}
\crefname{section}{Sec.}{Secs.}
\Crefname{section}{Section}{Sections}
\Crefname{table}{Table}{Tables}
\crefname{table}{Tab.}{Tabs.}

\def\cvprPaperID{*****} 
\def\confName{CVPR}
\def\confYear{2022}

\begin{document}

\title{VMFormer: End-to-End Video Matting with Transformer}

\author{ Jiachen Li\textsuperscript{1,3*}, Vidit Goel\textsuperscript{3}, Marianna Ohanyan\textsuperscript{3}, Shant Navasardyan\textsuperscript{3}, Yunchao Wei\textsuperscript{2}, Humphrey Shi\textsuperscript{1,3} \\
{\small \textsuperscript{1}SHI Lab $@$ University of Oregon \& UIUC, \textsuperscript{2}BJTU,   \textsuperscript{3}Picsart AI Research (PAIR)}\\
}

\twocolumn[{
\vspace{-3mm}
\maketitle
\begin{center}
    \vspace{-3mm}
    \includegraphics[width=1.0\textwidth]{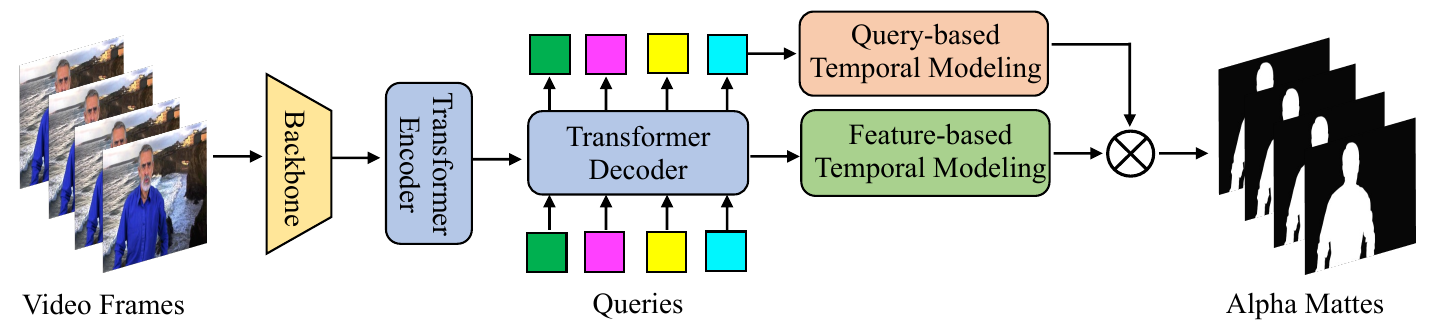}
    \captionof{figure}{
    \textbf{An overview of the VMFormer}. VMFormer has two branches for feature modeling and query modeling. The feature modeling branch takes video clips as input and extracts feature sequences through the backbone and the transformer encoder. The query modeling branch learns the global representation of feature sequences through the transformer decoder. Temporal modeling is applied to each branch individually to improve the temporal consistency of predictions. Finally, VMFormer makes predictions of alpha mattes with queries and feature maps. The detailed architecture of each branch is illustrated in Figure~\ref{fig:archi}.
    }
    \vspace{1mm}
    \label{fig:teaser}
\end{center}
}]

\let\thefootnote\relax\footnote{{*Work is partially done during an internship at Picsart AI Research}}

\begin{abstract}
    Video matting aims to predict the alpha mattes for each frame from a given input video sequence. Recent solutions to video matting have been dominated by deep convolutional neural networks~(CNN) for the past few years, which have become the \textit{de-facto} standard for academia and industry. 
    However, they have the inbuilt inductive bias of locality and do not capture the global characteristics of an image due to the CNN-based architectures. They also need long-range temporal modeling considering computational costs when dealing with feature maps of multiple frames. In this paper, we propose \textbf{VMFormer}: a transformer-based end-to-end method for video matting. It makes predictions on alpha mattes of each frame from learnable queries given a video input sequence.
    Specifically, it leverages self-attention layers to build global integration of feature sequences with short-range temporal modeling on successive frames. We further apply queries to learn global representations through cross-attention in the transformer decoder with long-range temporal modeling upon all queries. In the prediction stage, both queries and corresponding feature maps are used to make the final prediction of alpha matte. Experiments show that VMFormer outperforms previous CNN-based video matting methods on the composited benchmarks. To the best knowledge, it is the first end-to-end video matting solution built upon a full vision transformer with predictions on the learnable queries. The project is open-sourced at \href{https://chrisjuniorli.github.io/project/VMFormer/}{https://chrisjuniorli.github.io/project/VMFormer}.
\end{abstract}

\input{text/introduction}

\input{text/relatedworks}

\input{text/method}

\input{text/experiments}

\input{text/conclusion}

{\small
\bibliographystyle{ieee_fullname}
\bibliography{references}
}


\end{document}

%% file: text/introduction.tex
\section{Introduction}

Video matting aims to predict the alpha mattes of each frame from a given input video sequence. It has received considerable attention from industry and academia in recent years since it can be widely used in applications like video conferencing and video editing. Given a video sequence $\mathbf{I} = \{I_1, I_2, ... , I_T\}$, where each frame $I_i$ is assumed as a composition of a foreground image $F_i$ and a background image $B_i$~\cite{wang2008image} with a coefficient alpha matte map $\alpha_i \in [0,1]$:

\begin{equation}
 I_i = \alpha_i F_i + (1 - \alpha_i) B_i
\end{equation}

\noindent where  $\alpha_i \in [0,1]$ is the alpha matte. Video matting aims to predict precisely 
$\boldsymbol{\alpha} = \{\alpha_1, \alpha_2, ... , \alpha_T\}$. It is an under-constrained problem since for each pixel, there are seven unknown values (three from $F_i$, three from $B_i$, and one from $\alpha_i$) while only three known equations from $I_i$ on $R, G, B$ channels separately. To estimate $\boldsymbol{\alpha}$, some solutions expect an auxiliary input called \textit{trimap}, which is a segmentation ground truth of foreground, background, and unknown regions of frames. Such \textit{trimap-based} methods~\cite{xu2017deep,wang2018deep,zhu2017fast,li2020natural,yu2020high,sun2021deep,zhang2021attention} take inputs from both RGB frames and user-provided trimaps to predict alpha mattes based on deep convolutional neural networks~(CNN). Since trimap annotations are expensive for videos, some \textit{trimap-free} solutions~\cite{ke2020green,lin2021robust,sengupta2020background,lin2021real} are proposed for predicting alpha mattes using only RGB frames as inputs. 

However, most state-of-the-art CNN-based solutions suffer from the intrinsic inductive bias of CNN, especially on the locality of the receptive field and translation equivariance, which makes it hard to capture the global characteristics of the image while performing the task. Some methods~\cite{sun2021deep, Park_2022_CVPR} further adopt attention mechanisms to alleviate this issue while the decoder part remains CNN-based. Moreover, in these models, temporal modeling is usually performed on feature maps of successive frames~\cite{lin2021robust} or used as a post-processing approach to improving global temporal consistency~\cite{ke2020green}. These temporal modeling techniques lack long-range modeling and would be computationally expensive if implemented upon the feature maps of the whole video sequence. Recently, some vision transformer solutions~\cite{dosovitskiy2020image,liu2021swin} adopt self-attention model interaction between non-local pixels and get a global receptive field. Regarding downstream tasks like detection and segmentation, some methods~\cite{cheng2021per,wang2021max} leverage multiple queries and cross-attention to recognize all instances in images, which does not apply to video understanding since predictions are independent for each frame.

Motivated by these observations, we propose \textbf{VMFormer}: an end-to-end video matting solution based on a transformer as shown in Figure~\ref{fig:teaser}. It contains two branches: feature modeling and query modeling. The feature modeling branch has a CNN-based backbone and a transformer encoder for feature extraction of input video sequences. The self-attention layers in each transformer encoder block involve global modeling across the whole picture of each frame. The query modeling branch leverage the transformer decoder with cross-attention and queries for predictions of alpha mattes. Each query corresponds to an input frame. The cross-attention layer in each transformer decoder block integrates the learnable queries with corresponding feature sequences globally. In terms of temporal modeling, We add a short-range feature-based temporal modeling~(SFTM) module based on recurrent aggregations of successive feature maps. To add long-range temporal modeling in VMFormer while avoiding the heavy computations of applying it to feature sequences, we adopt a long-range query-based temporal model~(LQTM) that learns the weights for each query and updates all queries with the learned weights. In the end, the queries directly make matrix multiplications with the feature maps of the corresponding frame to predict alpha mattes accordingly.

We train and evaluate VMFormer with composited training and testing data from VideoMatte240K~\cite{lin2021real}, BG20K~\cite{li2021deep}, and DVM~\cite{sun2021deep}. We also evaluate VMFormer on the testing set used in RVM to compare with other state-of-the-art CNN-based models without re-training the model. Extensive experiments and ablation studies show that VMFormer performs better than previous CNN-based networks while keeping the real-time inference speed.

To summarize, our contributions are as follows:
\begin{itemize}
    \item We propose VMFormer: an end-to-end trimap-free video matting method based on a vision transformer and make predictions with learnable queries.
    
    \item We further build an efficient long-range query-based temporal modeling~(LQTM) module and short-range feature-based temporal modeling~(SFTM) module to improve the performance of VMFormer on video matting.
    
    \item To our best knowledge, VMFormer is the first end-to-end video matting solution built upon a transformer. We hope it motivates the community to explore more solutions on this track.
\end{itemize}

%% file: text/relatedworks.tex
\section{Related Works}
\subsection{Image Matting}
 Image matting is to estimate the accurate foreground in the image, which is an important computer vision task~\cite{wang2008image}. Mathematically, given an Image $I$ as a combination of unknown foreground image $F$ and background image $B$, which has 
\begin{equation}
     I = \alpha F + (1 - \alpha) B
\end{equation}
Previous image matting solutions mainly focus on low-level features to distinguish the transition areas between foregrounds and backgrounds~\cite{aksoy2017designing,chuang2001bayesian,bai2007geodesic,chen2013knn,grady2005random,feng2016cluster}. In recent years, deep learning based methods use neural networks to estimate alpha mattes from image feature maps with auxiliary trimap supervisions in an end-to-end manner~\cite{xu2017deep,wang2018deep,zhu2017fast,li2020natural,yu2020high,qiao2020attention, Park_2022_CVPR}. Other methods also explore trimap-free solutions with only segmentation network~\cite{chen2018semantic} or coarse annotations~\cite{liu2020boosting, yu2021mask}.

\begin{figure*}[tb]
\centering
\includegraphics[width=1.0\textwidth]{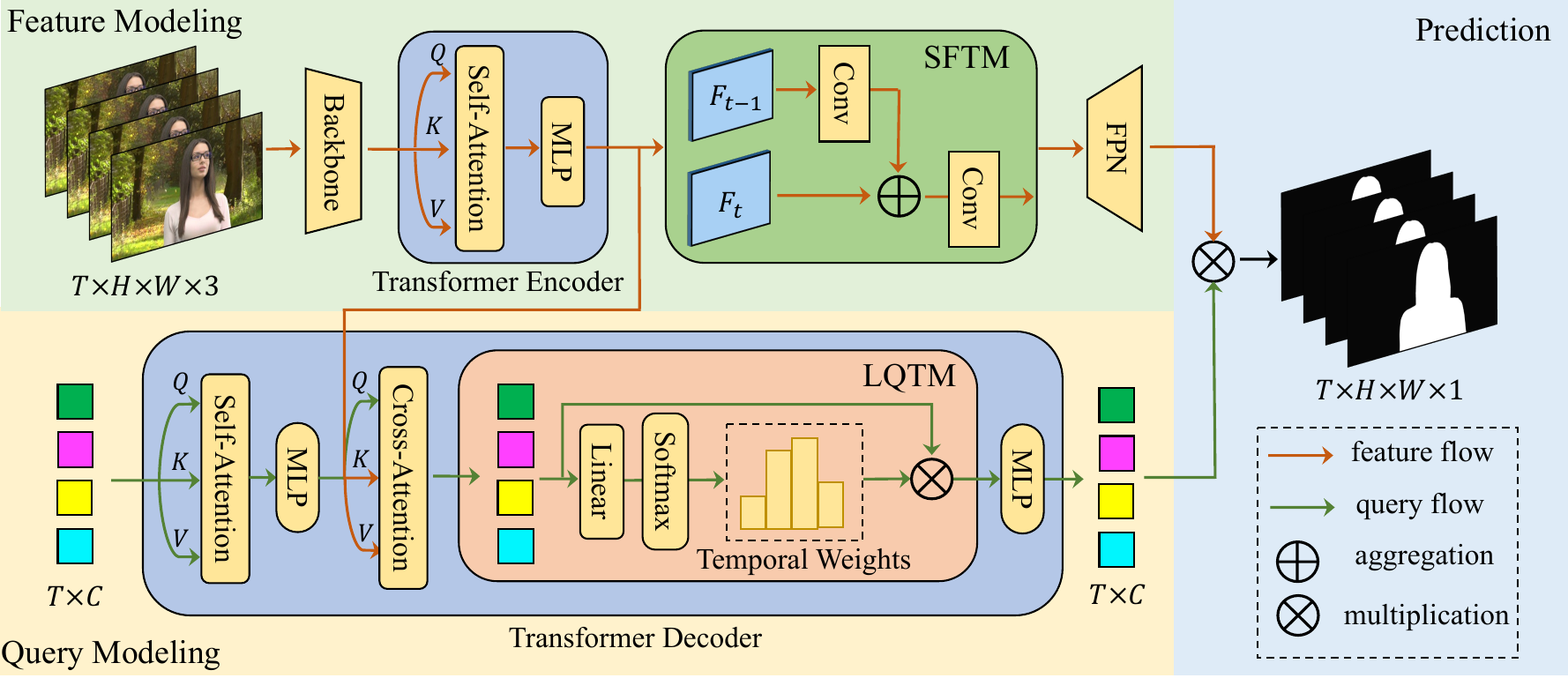}
\caption{\textbf{The Architecture of the VMFormer}. It contains two separate branches for modeling features and queries: a) The feature modeling path contains a CNN-based backbone network to extract feature pyramids and a transformer encoder that integrates feature sequences globally with short-range feature-based temporal modeling~(SFTM) on feature maps of successive frames. b) The query modeling path has a transformer decoder for queries to learn global representations of feature sequences. Long-range query-based temporal modeling~(LQTM) is built upon all queries with learnable temporal weights. The final alpha matte predictions are based on matrix multiplication between queries and feature maps. LayerNorm, residual connection, and repeated blocks are omitted for simplicity.}
\label{fig:archi}
\end{figure*}

\subsection{Video Matting}
When extending image matting to video sequences, trimap-based solutions focus on building temporal modeling with attention-based spatial-temporal feature aggregation~\cite{sun2021deep,zhang2021attention}. For trimap-free solutions, background matting series~\cite{sengupta2020background,lin2021real} use an auxiliary input of the background image for the first frame and estimate alpha mattes of the whole video sequences. MODNet~\cite{ke2020green} employs self-supervised method to model temporal consistency and RVM~\cite{lin2021robust} leverages segmentation data to make the matting model robust to real-world scenes. Although previous works explore the utilization of attention mechanisms in video matting solutions, the mainstream is still CNN-based models. Meanwhile, to our best knowledge, cross-attention has never been used so the predictions are still based on CNN kernels with limited perception. 

\subsection{Vision Transformer}
Recently, vision transformers are popular in the computer vision community. ViT~\cite{dosovitskiy2020image} and CCT~\cite{hassani2021escaping} use a vanilla transformer encoder for image classification. Swin~\cite{liu2021swin} and NAT~\cite{hassani2022neighborhood} further employ hierarchical local vision transformers as the backbone. DETR~\cite{carion2020end} adopts a full transformer for object detection and Deformable DETR~\cite{zhu2020deformable} proposes multi-scale deformable attention to reducing the training efforts with faster convergence. For pixel-prediction tasks like segmentation, Max-Deeplab~\cite{wang2021max}, MaskFormer~\cite{cheng2021per,cheng2021masked} and SeMask~\cite{jain2021semask} use learnable queries for masks predictions through cross-attention in image segmentation tasks. The queries are used for predicting instance-level segmentation results. In video instance segmentation, vision transformer based solutions~\cite{wang2021end,cheng2021mask2former,wu2021seqformer} further leverage queries for predictions of instance masks and build temporal matching across frames. However, there are not many explorations on utilizing a transformer to solve image-level video segmentation tasks like video matting, since predicting whole images across multiple frames brings a huge burden on memory, let alone the computationally expensive operators like self-attention and cross-attention. Furthermore, matching queries with feature sequences remains a challenging task for video understanding.

%% file: text/method.tex
\section{VMFormer}
In this section, we introduce how we establish the whole VMFormer architecture from feature modeling and query modeling to the prediction stage, as shown in Figure~\ref{fig:archi}.

\subsection{Feature Modeling}
\label{sec:feature}
\noindent \textbf{Backbone} The backbone network takes video sequence $\mathbf{I} = \{I_1, I_2, ... , I_T\} \in \mathbb{R}^{T \times H \times W \times 3}$ as input where $T$ is the number of frames and extracts a pyramid of feature maps $\mathbf{F} = \{F_1, F_2, F_3, F_4\}$ with dimension $F_i \in  \mathbb{R}^{T \times \frac{H}{2^i} \times \frac{W}{2^i} \times C_i}$. Then, they are projected to same dimension $C$ and flattened into feature sequences $\mathbf{X} \in \mathbb{R}^{T \times L \times C} $ where $L = \sum_{i=1}^4 \frac{HW}{4^{i}}$ as inputs for transformer.

\noindent \textbf{Transformer Encoder} The transformer encoder is built upon $N_{enc}$ consecutive blocks. Each encoder block consists of a self-attention layer~(SA) and a multi-layer perceptron~(MLP). LayerNorm~(LN) and residual connection are applied to the output of them. Given input feature sequences $\mathbf{X}$, for each frame $X_i \in \mathbb{R}^{L \times C} $
\begin{equation}
    X'_{i} = LN(SA(X_{i}) + X_{i})
\end{equation}
\vspace{-5mm}
\begin{equation}
    X_{enc_i} = LN(MLP(X'_{i}) + X'_{i})
\end{equation}

\noindent The output $X_{enc_i} \in \mathbb{R}^{L \times C}$ is denoted as feature sequence of each frame. The overall feature sequences are the concatenation of all frames as $\mathbf{X}_{enc} = \{X_{enc_1}, X_{enc_2}, ..., X_{enc_T}\} \in \mathbb{R}^{T \times L \times C}$.

\noindent \textbf{SFTM} We reshape the feature sequence $\mathbf{X}_{enc}$ into feature pyramids $\mathbf{F}_{enc} = \{F_{enc_1}, F_{enc_2}, F_{enc_3}, F_{enc_4}\}$ for predictions. To involve temporal modeling between feature maps of consecutive frames, we first try a simple feature aggregation that aggregates feature maps of two consecutive frames. Considering $F^{t}_{enc_i}$ and $F^{t-1}_{enc_i}$ as feature maps of frame t and t-1, feature aggregation is introduced as 
\begin{equation}
    F^{'t}_{enc_i} = Conv(F^t_{enc_{i}}+F^{t-1}_{enc_i})
\end{equation}
It brings some performance improvement as shown in the ablation study~\ref{sec:ablation} of
short-range feature-based temporal modeling~(SFTM). To make the temporal context transferable across frames explicitly, we further propose a recurrent aggregation, which updates the feature maps at each frame with the recurrent features $h^t_{enc_i}$ iteratively by
\begin{equation}
   F^{'t}_{enc_i} = Conv(F^t_{enc_{i}} + h^{t-1}_{enc_i})
\end{equation}
the $h^t_{enc_i}$ is initialized with $F^0_{enc_i}$ and updated the same way as $F^{'t}_{enc_i}$. The experiments show that the model benefits from recurrent aggregation more than direct feature aggregation, and we use recurrent aggregation as the SFTM module in the VMFormer. Then, we use an FPN on top of these updated feature pyramids by
\begin{equation}
    F''_i = F'_{enc_i} + Upsample(Conv(F'_{enc_{i+1}}))
\end{equation}

\subsection{Query Modeling}
\label{sec:query}
\noindent \textbf{Transformer Decoder} For the query modeling part, the main goal is to use the cross-attention module to make learnable queries interact with whole feature sequences so that they can learn global representations and make better predictions for alpha mattes. To make training and inference flexible, we first generate a sequence of queries $\mathbf{Q} \in \mathbb{R}^{T \times C}$ as inputs to $N_{dec}$ consecutive decoder blocks, which has the same length $T$ to the feature sequences so that each query can interact with corresponding feature sequence $X_{enc_i}$. In each decoder block, each query $Q_i$ firstly goes through the self-attention layer~(SA) based module,
\begin{equation}
    Q'_{i} = LN(SA(Q_{i}) + Q_{i})
\end{equation}
\vspace{-5mm}
\begin{equation}
    Q_{i} = LN(MLP(Q'_{i}) + Q'_{i})
\end{equation}
Then, it interacts with the corresponding feature sequence $X_{enc_i}$ globally in the cross-attention layer~(CA) based module. In this module, the key $K_{enc_i}$ and value $V_{enc_i}$ are linear transformation from feature sequence $X_{enc_i}$,
\begin{equation}
   CA(Q_i, X_{enc_i}) = softmax(Q_{i} K_{enc_i}^T) \times V_{enc_i}
\end{equation}

\noindent \textbf{LQTM} To add temporal modeling based on queries since they have learned effective global representation from the cross-attention layer. They are less computationally expensive than feature maps. We add an attention-based long-range temporal modeling~(LQTM), and the attention weights are learned from the queries themselves. We first apply a self-attention-based approach, which uses a multi-head self-attention layer to compute the $Q_{tem}$ with temporal attention weights of all queries,
\begin{equation}
     Q_{tem} = SA(CA(Q_i, X_{enc_i}))
\end{equation}
where the learnable queries show better time consistent with the temporal attention weights, as shown in ablation study~\ref{sec:ablation} of long-range query-based temporal modeling~(LQTM). Considering the simplicity of queries and self-attention may be overkill for queries, we further use additive attention, which adopts a normalized linear layer to learn the temporal weights 
\begin{equation}
    W_i = softmax(linear(CA(Q_i, X_{enc_i})))
\end{equation}
\vspace{-5mm}
\begin{equation}
    Q_{tem} = \sum_{i=1}^T W_i \times Q_i
\end{equation}
It is also computationally efficient, which only has a cost of $\mathcal{O}(TC)$ that is much less than $\mathcal{O}(THWC)$ if it is applied to video feature sequences. LayerNorm, residual connection, and MLP are added to $Q_i$ followingly,
\begin{equation}
    Q'_i = LN(CA(Q_{i}, X_{enc_i}) + Q_{i} + Q_{tem})
\end{equation}
\vspace{-5mm}
\begin{equation}
    Q_{dec_i} = LN(MLP(Q'_{i}) + Q'_{i})
\end{equation}
Both the outputs of the learnable queries and the feature sequences at each frame $Q_{dec_i} \in \mathbb{R}^{C}$ and $X_{enc_i} \in \mathbb{R}^{L \times C}$ (Since $X_{enc_i}$ stays the same after cross-attention, so the notation remains unchanged) are concatenated as $\mathbf{Q}_{dec} \in \mathbb{R}^{T \times C}$ and $\mathbf{X}_{enc} \in \mathbb{R}^{T \times L \times C}$ for final predictions.

\subsection{Prediction Stage}
\noindent \textbf{Predictor} After getting queries $\mathbf{Q}_{dec}$ from the transformer decoder, we make predictions based on the largest feature map $F''_1$ and the queries $\mathbf{Q}_{dec}$, which are projected to $F''_1$ with a batch matrix multiplication
\begin{equation}
   \boldsymbol{\alpha}  =  F''_1 \otimes \mathbf{Q}_{dec}
\end{equation}
where $\otimes$ denotes batch matrix multiplication, and the outputs are further upsampled as final predictions of alpha mattes $\boldsymbol{\alpha} \in \mathbb{R}^{T \times H \times W \times 1}$.

\noindent \textbf{Loss Functions}
To compute losses during the training stage, we use a combination loss function
\begin{equation}
    L = \lambda L_{focal} + L_{dice} + L_{tmp}
\end{equation}
which includes pixel-level focal loss~\cite{lin2017focal} with balanced coefficient $\lambda$, $\alpha$ is a balanced term and $p_{x,y}$ is the prediction on the pixel with coordinate $(x,y)$.
\begin{equation}
    L _{focal} = - \alpha (1 - p_{x,y})^{\gamma}\log(p_{x,y})
\end{equation}
$L_{dice}$ is dice loss~\cite{milletari2016v} and $g_{x,y}$ is the ground truth on the pixel with coordinate $(x,y)$. $L_{tmp}$ is a temporal consistency loss~\cite{lin2021robust}. 
\begin{equation}
    L _{dice} = 1 - \frac{2 \times \sum_{x,y} p_{x,y} g_{x,y}}{\sum_{x,y} p^2_{x,y} + \sum_{x,y} g^2_{x,y}}
\end{equation}
\begin{equation}
    L _{tmp} =  \| \frac{\partial p_{x,y}}{\partial t} - \frac{\partial \hat g_{x,y}}{\partial t} \|_2 \
\end{equation}

\input{tables/ablation_encdec}
\input{tables/ablation_coefficient}
\input{tables/ablation_supervision}
\noindent \textbf{Inference} During inference process, given a input video clip $\mathbf{I} \in \mathbb{R}^{t \times H \times W \times 3}$ and $t$ is the length of the video clip. It is split into consecutive sequences $\mathbf{I}_i \in \mathbb{R}^{T \times H \times W \times 3}$ and $T$ could be any length during inference. Then, each $\mathbf{I}_i$ is sent to VMFormer from the backbone, transformer encoder, and transformer decoder to predictor sequentially. We only use the outputs $\boldsymbol{\alpha}_i \in \mathbb{R}^{T \times H \times W \times 1}$ from the last transformer decoder block as predictions. Finally, all outputs $\boldsymbol{\alpha}_i$ are combined to $\boldsymbol{\alpha} \in \mathbb{R}^{t \times H \times W \times 1}$ as alpha matte predictions of the corresponding input video clip $\mathbf{I}$.

%% file: tables/ablation_encdec.tex
\begin{table}[tb]
\centering
\begin{tabular}{c|ccccc}
  &  MAD$\downarrow$ &  MSE$\downarrow$ &  Grad$\downarrow$ &  Conn$\downarrow$  \\ \hline
  $N_{enc}=1$ &7.02 &2.01 &1.38 &0.56 \\ 
  $N_{enc}=2$  &\textbf{6.57} &\textbf{1.57} &\textbf{1.10} &\textbf{0.49} \\ 
   $N_{dec}=1$ &7.02 &2.01 &1.38 &0.56 \\ 
 $N_{dec}=2$ &6.90 &1.93 &1.36 &0.54 \\
\end{tabular}
\vspace{-1mm}
\caption{Ablation Study on Number of Encoder and Decoder Blocks. \textbf{Bold} indicates the best performance among these models. The model
benefits from deeper architecture like other CNN-based models.}
\vspace{-1mm}
\label{tab:ablation_encdec}
\end{table}

%% file: tables/ablation_coefficient.tex
\begin{table}[tb]
\centering
\begin{tabular}{c|ccccc}
  & MAD$\downarrow$ &  MSE$\downarrow$ &  Grad$\downarrow$ &  Conn$\downarrow$ \\ \hline
 $\lambda$ = 1 &7.02 &2.01 &1.38 &0.56 \\
$\lambda$ =5 &\textbf{7.02} &1.81 &1.25 &\textbf{0.56} \\
$\lambda$ = 10 &7.17 &\textbf{1.77} &\textbf{1.22} &0.57 \\
\end{tabular}
\vspace{-1mm}
\caption{Ablation Study on the Coefficient $\lambda$ of Focal Loss. \textbf{Bold} indicates the best performance among these models. We chose $\lambda = 5$ for the final model.}
\vspace{-1mm}
\label{tab:ablation_coefficient}
\end{table}

%% file: tables/ablation_supervision.tex
\begin{table}[tb]
\centering
\begin{tabular}{c|cccc}
  & MAD$\downarrow$ &  MSE$\downarrow$ &  Grad$\downarrow$ &  Conn$\downarrow$   \\ \hline
 $F_1$ &\textbf{7.02} &\textbf{2.01} &1.38 &\textbf{0.56} \\
 $F_1 + F_2$ &7.03 &2.04   &1.38 &0.56 \\ 
  $F_1 + F_2 + F_3$ &7.02 &2.01   &\textbf{1.35} &0.56 \\
\end{tabular}
\vspace{-1mm}
\caption{Ablation Study on the Auxiliary Supervision. \textbf{Bold} indicates the best performance among these models. We only select supervision from the final predictions $F_1$.}
\vspace{-1mm}
\label{tab:ablation_super}
\end{table}

%% file: text/experiments.tex
\section{Experiments}
\subsection{Dataset and Evaluation}
\noindent \textbf{Dataset} Following previous works, we also use composited training data from the merged image foreground dataset ImageMatte~\cite{lin2021robust}, the video foreground dataset VideoMatte240K~\cite{lin2021real}, the image background dataset BG20K~\cite{li2021deep} and the video background dataset from DVM~\cite{sun2021deep}. Following RVM~\cite{lin2021robust}, we split VideoMatte240K into 475/4/5 video clips for training, validating, and testing. Similarly, BG20K is split into 15000/500/4500 image sets, and DVM is split into 3080/37/162 image sets for training and testing. During training, we randomly select foreground from ImageMatte and VideoMatte240K, background from BG20K, and DVM to composite video clips on the fly. During testing, we composite 50 challenging clips of resolutions at $512 \times 288$ and $1920 \times 1080$ for testing the VMFormer and other models.

\input{tables/ablation_scheduler}
\input{tables/ablation_temporal}

\noindent \textbf{Evaluation} We mainly consider video matting accuracy and temporal coherence for evaluation. For video matting accuracy, we use Mean Absolute Difference~(MAD), Mean Squared Error~(MSE), Gradient~(Grad), and Connectivity~(Conn)~\cite{rhemann2009perceptually} as evaluation metrics. We also scale MAD, MSE, Grad, and Conn by $10^3$, $10^3$, $10^{-3}$, and $10^{-3}$, respectively, for convenience of reference. For all these metrics, the lower number represents better performance.

\subsection{Implementation Details}
\noindent \textbf{Training Settings} For training the VMFormer network, we use 8 RTX A6000 GPUs with batch size at 2 video clips per GPU. The optimizer is Adam, with different initial learning rates at different model modules. The backbone is under $2 \times 10^{-5}$ and other modules are under $2 \times 10^{-4}$ with weight decay at $1 \times 10^{-4}$. All models in ablation studies are trained for 12 epochs and the learning rates are decayed at the 6th and 8th epoch by a factor of 0.1. Given an input video clip, it is further split into sequences with length $T=5$ for training and inference due to the GPU memory constraints. Then, for augmentation, each frame of the input video sequences is randomly resized to the longest side in the range of {[288, 320, 352, 392, 416, 448, 480, 512]} with random horizontal flipping.

\noindent \textbf{Backbone} The backbone we used for VMFormer is ImageNet~\cite{krizhevsky2012imagenet} pretrained Mobilenetv3-Large~\cite{howard2019searching} since it is the most frequently used backbones in prior works. We use the implementation and weights from $\mathtt{torchvision}$~\cite{paszke2019pytorch}.

\input{tables/sota_ours_public}

\noindent \textbf{Transformer Encoder} The transformer encoder and decoder are based on Deformable DETR~\cite{zhu2020deformable}. We adopt multi-scale deformable self-attention to replace multi-head attention for faster convergence and efficient computing. We set the channel $C=256$, the number of attention head $M=8$, and the number of key sampling points $K=4$ for each multi-scale deformable attention module.

\noindent \textbf{Transformer Decoder} The number of queries is set to be 5 for training at the transformer decoder with channels $C=256$ to be compatible with the length $T=5$ of training samples. The positional embedding has the same dimensions as the queries. During feed-forward propagation, both the self-attention and cross-attention are implemented based on a multi-scale deformable attention module, in which the number of attention head $M=8$ and the number of key sampling points $K=4$.

\noindent \textbf{Predictor} After getting learnable queries and the recovered feature maps from the feature pyramid network, we simply multiply queries to feature maps with a batch matrix multiplication implemented based on $\mathtt{torch.einsum}$~\cite{rogozhnikov2021einops}. Then, the predicted alpha mattes are upsampled by bilinear interpolation to the original size of input sequences. 

\noindent \textbf{Loss Functions} For focal loss, we set $\lambda$ to 5 for final models based on ablation experiments on $\lambda$. $\alpha$ and $\gamma$ are set to $0.25$ and $2$, respectively.

\noindent \textbf{Inference} During inference, each video clip for testing has 100 frames, and we set $T=20$ for inference with input at $512 \times 288$ and $T=5$ for inference with input at $1920 \times 1080$ on a single RTX A6000 GPU with batch size at 1.

\input{tables/sota_rvm_public.tex}
\input{tables/speed}

 \begin{figure*}[tb]
\centering
\includegraphics[width=1.0\textwidth]{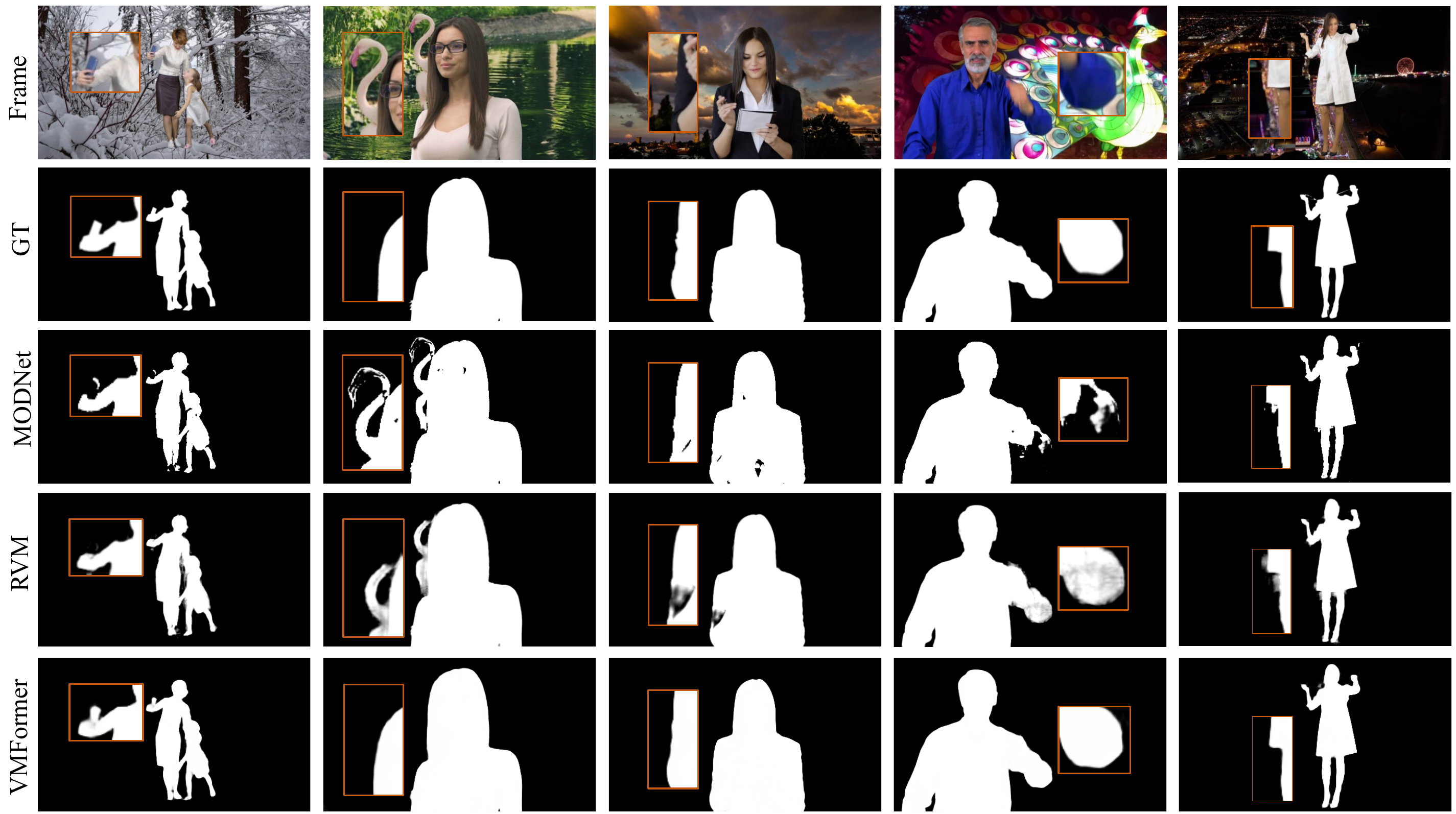}
\caption{Visualization of alpha matte predictions from MODNet, RVM, and VMFormer under challenging frames from the composited test set. VMFormer shows a better ability to distinguish ambiguous foreground from background regions as shown in the magnified image patches. Please zoom in for details.}
\label{fig:vis}
\end{figure*}

\subsection{Ablation Studies}
\label{sec:ablation}

\noindent \textbf{Number of Encoder and Decoder Blocks} In Table~\ref{tab:ablation_encdec}, we change the number of transformer encoder and decoder blocks to evaluate if deeper architecture improves the performance. We set $N_{enc}=1, 2$ and $N_{dec}=1, 2$ due to the GPU memory constraints. It shows that the model benefits from deeper architecture like other CNN-based models.

\noindent \textbf{Coefficient of Focal Loss} Since we observe that the loss value of the focal loss is much smaller than dice loss and temporal consistency loss, we add a balanced coefficient $\lambda$ for focal loss and test the performance under different values of $\lambda$. It shows that improving the coefficient of focal loss could improve temporal consistency as shown in Table~\ref{tab:ablation_coefficient} and we choose $\lambda=5$ for the VMFormer and following experiments.

\noindent \textbf{Training Epochs} In all ablation studies above, we use a 12-epochs training setting for fair comparisons. We extend the training scheduler to 24 and 36 epochs to evaluate the effect of a longer training scheduler, considering that recent vision transformer works~\cite{carion2020end,zhu2020deformable} require longer training schedulers for better convergence. The learning rates decayed at the 12th/18th and 20th/30th epoch by 0.1, respectively. Experiments show that a longer training scheduler consistently improves the performance of VMFormer as shown in Table~\ref{tab:ablation_scheduler}.

\noindent \textbf{Auxiliary Supervisions} Considering that queries could make predictions based on each level of the feature pyramids, we further compute losses on predictions of the lower resolution feature maps $F_2$ and $F_3$. Experiment results are shown in Table~\ref{tab:ablation_super} that extra supervisions from $F_2$ and $F_3$ are not improving performance. To summarize, both adding auxiliary supervisions from more transformer decoder blocks and predictions of intermediate levels are not helpful to VMFormer. This conclusion is not aligned with observations from other vision transformers~\cite{carion2020end}\cite{wu2021seqformer}, we think it implies that prediction of a single-channel alpha matte is relatively easy for transformer compared with instance-level recognition tasks.

\noindent \textbf{Temporal Modeling} To evaluate the effectiveness of temporal modeling module SFTM and LQTM, we train VMFormer without any temporal modeling in $3\times$ scheduler as a strong baseline for comparison. For SFTM, we evaluate both feature aggregation and recurrent aggregation as discussed in Section~\ref{sec:feature}, and recurrent aggregation shows better performance on all metrics compared to feature aggregation. We further evaluate self-attention and additive attention for LQTM as discussed in Section~\ref{sec:query}, and additive attention achieves better performance with lower computations. Ultimately, we adopt recurrent aggregation-based SFTM and additive attention-based LQTM for the VMFormer. Experiments show that the model benefits from both sides and achieves the best performance, as shown in Table~\ref{tab:ablation_temporal}.

\subsection{Comparisons to State-of-the-art Methods}
To make fair comparisons to previous methods, we select the most recent state-of-the-art trimap-free video matting solutions, including BGMv2~\cite{lin2021real}, MODNet~\cite{sun2021modnet} and RVM~\cite{lin2021robust} for comparisons, and evaluate all models on the same composited testing set. Then, we evaluate the pre-trained VMFormer on the test set used in RVM~\cite{lin2021robust}.

\noindent \textbf{Composited Test Set}
In Table~\ref{tab:ours_public}, we compare VMFormer to three state-of-the-art CNN-based trimap-free video matting solutions BGMv2~\cite{lin2021real}, MODNet~\cite{sun2021modnet} and RVM~\cite{lin2021robust}. We used the pre-trained MobileNetV3~\cite{howard2019searching} based checkpoints since they perform better than the reproduced ones that re-trained on our training sets. We evaluate all models on the composited test set that contains 50 challenging video clips, and each one has 100 frames for evaluation. They are tested under the resolution at $512 \times 288$ and $1920 \times 1080$ to show the robustness of the models to different resolutions. It shows that VMFormer reaches the best performance on all metrics concerning to the accuracy of alpha matte predictions among all these models. Specifically, it reaches 4.91 MAD, 0.55 MSE, 0.40 Grad, and 0.25 Conn under both inputs of $512 \times 288$. For input of $1920 \times 1080$, it reaches 4.81 MAD, 0.78 MSE, 4.90 Grad, and 3.34 Conn, respectively.

\noindent \textbf{RVM's Test Set}
In Table~\ref{tab:rvm_public}, we further compare VMFormer to other methods on the composited test set used in RVM. All these methods except VMFormer are trained on both matting data and segmentation data~\cite{lin2014microsoft, spd2018, yang2019video}. We evaluate the models under input resolution both at $512 \times 288$ and $1920 \times 1080$ to show the robustness of the models to different resolutions.
It shows VMFormer outperforms all other models without further joint training on the segmentation data under both inputs of low resolution and high resolution.
 
\noindent \textbf{Speed Comparison}
Since video matting is close to industrial applications, thus real-time inference is always preferable for video matting algorithms. Previous CNN-based methods MODNet~\cite{sun2021modnet}, BGMv2~\cite{lin2021real} and RVM~\cite{lin2021robust} have shown real-time inference speed in Table~\ref{tab:speed}. All these methods are tested on a single A6000 GPU under the input of $512 \times 288$ without extra optimization during inference. It shows VMFormer could reach real-time inference and comparable FPS to CNN-based models.

\begin{figure}[tb]
\centering
\includegraphics[width=0.5\textwidth]{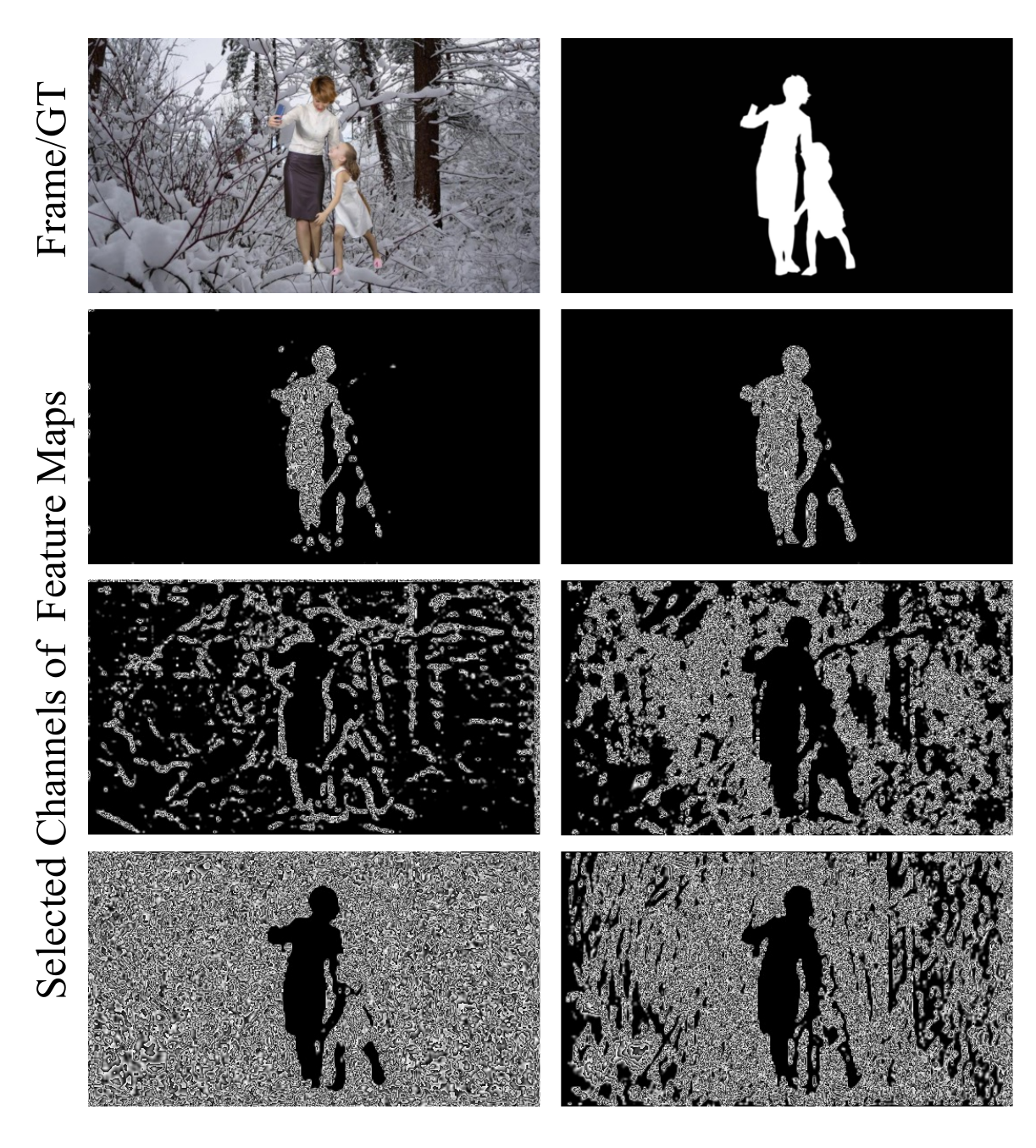}
\vspace{-2mm}
\caption{Visualization of the largest feature map ($\frac{H}{2} \times \frac{W}{2}$) from VMFormer under a challenging input frame. We selected feature maps of 6 single channels for presentation. VMFormer can better capture global information and distinguish foreground/background in the whole image. Please zoom in for details.}
\label{fig:vis_feature}
\vspace{-5mm}
\end{figure}

\subsection{Visualization}
\noindent \textbf{Challenging Video Frames} We visualize some challenging composited video frames from the test set and the corresponding alpha matte predictions as shown in Figure~\ref{fig:vis}.  It shows that VMFormer can better distinguish ambiguous foregrounds from backgrounds compared with other CNN-based models like MODNet and RVM, especially on color obfuscation.

\noindent \textbf{Selected Channels of Feature Maps } We further visualize some channels of feature maps in Figure~\ref{fig:vis_feature} and we observe that VMFormer has many redundant empty predictions, which is aligned with observations from other vision transformer works~\cite{yuan2021tokens}. Moreover, VMFormer generates high-quality feature maps that are useful for accurate alpha matte predictions.

%% file: tables/ablation_scheduler.tex
\begin{table}[tb]
\centering
\begin{tabular}{c|ccccc}
  &  MAD$\downarrow$ &  MSE$\downarrow$ &  Grad$\downarrow$ &  Conn$\downarrow$  \\ \hline
  12-epoch &7.02 &2.01 &1.38 &0.56 \\ 
  24-epoch &6.40 &1.44 &0.96 &0.47 \\ 
  36-epoch &\textbf{6.24} &\textbf{1.30} &\textbf{0.91} &\textbf{0.44} \\
\end{tabular}
\vspace{-1mm}
\caption{Ablation Study on the Training Epochs. \textbf{Bold} indicates the best performance among these models. The model benefits from longer training like other vision transformer models. We use the result of 36-epoch training as a strong baseline model.}
\vspace{-1mm}
\label{tab:ablation_scheduler}
\end{table}

%% file: tables/ablation_temporal.tex
\begin{table}[tb]
\centering
\resizebox{0.5\textwidth}{!}{
\begin{tabular}{c|ccccc}
  & MAD$\downarrow$ &  MSE$\downarrow$ &  Grad$\downarrow$ &  Conn$\downarrow$ \\ \hline 
Baseline &6.24 &1.30 &0.91 &0.44 \\  \hline
\multicolumn{1}{l}{\textit{SFTM}} \\ \hline
+ feature aggregate  &6.06 &1.28 &0.82 &0.42 \\ 
+ recurrent aggregate &5.14 &0.71 &0.53 &0.29 \\ \hline
\multicolumn{1}{l}{\textit{LQTM}} \\ \hline
+ self attention &5.67 &0.79 &0.62 &0.39 \\
+ additive attention &5.56 &0.75  &0.61 &0.33 \\ \hline
\multicolumn{1}{l}{\textit{Both}} &\textbf{4.91} &\textbf{0.55} &\textbf{0.40} &\textbf{0.25} \\ \hline
\end{tabular}}
\vspace{-1mm}
\caption{Ablation Study on the Short-range Feature-based Temporal Modeling~(SFTM) and Long-range Query-based Temporal Modeling~(LQTM) Modules. We use recurrent aggregate for SFTM and additive attention for LQTM in the final model of VMFormer.}
\vspace{-1mm}
\label{tab:ablation_temporal}
\end{table}

%% file: tables/sota_ours_public.tex
\begin{table}[tb]
\centering
\resizebox{0.5\textwidth}{!}{
\begin{tabular}{c|cccc}
Model  &  MAD$\downarrow$ &  MSE$\downarrow$ &  Grad$\downarrow$ & Conn$\downarrow$ \\ \hline \hline 
\multicolumn{1}{l}{\textit{LR: 512 $\times$ 288}} \\ \hline \hline 
BGMv2$\dagger$~\cite{lin2021real} &6.63 &1.79 &1.54 &0.50 \\
MODNet~\cite{sun2021modnet} &10.39 &5.65  &2.02 &1.04 \\
RVM~\cite{lin2021robust} &5.99 &1.17 &1.10 &0.34 \\ \hline
VMFormer &\textbf{4.91} &\textbf{0.55} &\textbf{0.40} &\textbf{0.25}\\ \hline \hline 
\multicolumn{1}{l}{\textit{HD: 1920 $\times$ 1080}} \\ \hline \hline 
BGMv2$\dagger$~\cite{lin2021real} &27.01 &21.31 &20.34 &49.69 \\ 
MODNet~\cite{sun2021modnet} &11.56 &6.47  &15.23 &17.23 \\
RVM~\cite{lin2021robust} &6.45 &1.63 &11.59 &5.74 \\ \hline
VMFormer &\textbf{4.81} &\textbf{0.78} &\textbf{4.90} &\textbf{3.34} \\ \hline
\end{tabular}}
\caption{Comparisons on the Composited Test Set. \textbf{Bold} indicates the best performance among these models under the inputs with the same resolution. We tested the data with inputs at low-resolution $512 \times 288$ and high-resolution $1920 \times 1080$. $\dagger$ indicates an augmented BGMv2 with background input in each frame.}
\label{tab:ours_public}
\end{table}

%% file: tables/sota_rvm_public.tex
\begin{table}[tb]
\centering
\resizebox{0.5\textwidth}{!}{
\begin{tabular}{c|cccc}
Model &  MAD$\downarrow$ &  MSE$\downarrow$ &  Grad$\downarrow$ &  Conn$\downarrow$ \\ \hline  \hline  
\multicolumn{1}{l}{\textit{LR: 512 $\times$ 288}} \\ \hline  \hline 
DeepLabV3~\cite{chen2017rethinking} &14.47 &9.67 &8.55 &1.69 \\
FBA~\cite{forte2020f} &8.36 &3.37 &2.09 &0.75 \\
BGMv2~\cite{lin2021real} &25.19 &19.63 &2.28 &3.26 \\ 
MODNet~\cite{sun2021modnet} &9.41 &4.30 &1.89 &0.81 \\
RVM~\cite{lin2021robust} &6.08 &1.47 &0.88 &0.41 \\ \hline
VMFormer &\textbf{6.02} &\textbf{1.00} &\textbf{0.75} &\textbf{0.37} \\ \hline   \hline
\multicolumn{1}{l}{\textit{HD: 1920 $\times$ 1080}} \\ \hline \hline 
BGMv2$\dagger$~\cite{sun2021modnet} &20.35 &14.26 &22.79 &- \\
MODNet~\cite{sun2021modnet} &11.13 &5.54 &15.30 &- \\ 
RVM~\cite{lin2021robust} &6.57 &1.93 &10.55 &- \\ \hline
VMFormer &\textbf{6.20} &\textbf{1.53} &\textbf{6.30} &\textbf{5.38} \\ \hline 
\end{tabular}}
\caption{Comparisons on the Composited Test Set from RVM. \textbf{Bold} indicates the best performance among these models under the inputs with the same resolution. We tested the data with inputs at low-resolution $512 \times 288$ and high-resolution $1920 \times 1080$ following practice in RVM. Conn under $1920 \times 1080$ were not reported in RVM's paper due to expensive computation. $\dagger$BGMv2 was tested by us with an augmented version since it was not reported in RVM. VMFormer shows state-of-the-art performance without re-training the model.}
\label{tab:rvm_public}
\end{table}

%% file: tables/speed.tex
\begin{table}[tb]
\centering
\begin{tabular}{c|ccc}
 Model &  Frames &  Input Resolution &  FPS   \\ \hline
  BGMv2~\cite{lin2021real} &1000 &$512 \times 288$ &195.9  \\
MODNet~\cite{lin2021real} &1000 &$512 \times 288$  &124.0  \\
RVM~\cite{lin2021robust}  &1000 &$512 \times 288$ &131.5  \\ \hline
VMFormer   &1000 &$512 \times 288$ & 109.5 \\ 
\end{tabular}
\vspace{-1mm}
\caption{Speed Comparisons between different models. All models are tested on a single A6000 GPU without extra optimization on inference.}
\vspace{-1mm}
\label{tab:speed}
\end{table}

%% file: text/conclusion.tex
\section{Conclusion}
In this paper, we propose VMFormer: an end-to-end transformer-based method for video matting, which uses self-attention in the transformer encoder to model the global integration of feature sequences. It leverages cross-attention in the transformer decoder to model interactions between whole feature sequences and queries as query modeling. We further build a long-range query-based temporal modeling~(LQTM) module and a short-range feature-based temporal modeling~(SFTM) module to encode temporal information for better predictions of alpha mattes. Extensive experiments show that VMFormer outperforms previous CNN-based methods and has robust performance on other benchmarks with comparable real-time inference speeds. 